\documentclass[journal]{IEEEtran}
\usepackage{lipsum,graphicx}

\usepackage{array}
\usepackage{caption}
\usepackage{subcaption}
\usepackage{capt-of}
\usepackage{multirow}
\usepackage{makecell} 
\setcellgapes{5pt}
\usepackage{tikz}
\usepackage{subfloat}
\usepackage{hyperref}
\usepackage{tabularx}
\usepackage{colortbl}
\usepackage{hhline}
\usepackage[utf8]{inputenc}
\usepackage[switch]{lineno} 
\usepackage[font=small]{caption}
\usepackage[inline]{enumitem}
\usepackage{threeparttable}

\begin{document}
\title{Point Transformer for Shape Classification and Retrieval of 3D and Urban Roof Point Clouds}

\author{Dimple~A~Shajahan,
 Mukund~Varma~T
and~Ramanathan~Muthuganapathy\vspace{-4ex}
\thanks{Manuscript received 16 June '20. The authors are with IIT Madras, Chennai-36, India. (email: dimpleshaj@gmail.com, mukundvarmat@gmail.com, emry01@gmail.com)}}

\maketitle

\begin{abstract} 
The success of deep learning methods led to significant breakthroughs in 3-D point cloud processing tasks with applications in remote sensing. Existing methods utilize convolutions that have some limitations, as they assume a uniform input distribution and cannot learn long-range dependencies. Recent works have shown that adding attention in conjunction with these methods improves performance. This raises a question: can attention layers completely replace convolutions? This paper proposes a fully attentional model - {\em Point Transformer}, for deriving a rich point cloud representation. The model's shape classification and retrieval performance are evaluated on a large-scale urban dataset - RoofN3D and a standard benchmark dataset ModelNet40. Extensive experiments are conducted to test the model's robustness to unseen point corruptions for analyzing its effectiveness on real datasets. The proposed method outperforms other state-of-the-art models in the RoofN3D dataset, gives competitive results in the ModelNet40 benchmark, and showcases high robustness to various unseen point corruptions. Furthermore, the model is highly memory and space efficient when compared to other methods.
\end{abstract}

\section{Introduction}
\IEEEPARstart{T}{he}
shape classification and retrieval of 3D point clouds is an important task in computer graphics and has a lot of applications in remote sensing. Building roof point clouds are very significant in urban modeling and improved Airborne Laser Scanning (ALS) techniques have made it easier to rapidly obtain the three-dimensional (3-D) large scale urban scene.  However, the captured point clouds are sparse, noisy, and incomplete, and there is a vital requirement for robustness to these issues. Recent improvements in deep learning techniques and the emergence of 3-D shape datasets such as ModelNet \cite{ModelNet40}, ShapeNet \cite{ShapeNet} has led to advancements in 3-D shape classification and retrieval. However, deep learning is still not widely explored in remote sensing and Geographic Information System (GIS). 

Deep learning approaches related to 3-D shape representation can be roughly categorized into view-based, voxel-based, and point-based methods. In view-based techniques \cite{mvcnn}, multiple views of the point cloud are generated and fed to a dense convolutional neural network (CNN) to derive feature representations. The performance of these models is limited by the number of views, high storage, and computational requirements.
Voxel-based techniques \cite{Voxel} use 3-D CNNs and convert points to voxels, which leads to loss of intrinsic geometric properties of the point cloud. On the other hand, point-based techniques \cite{PointNet, PointNet++, PointCNN, SONet, DensePoint} directly utilize point-wise information and are increasingly becoming popular these days. Point clouds are non-uniform, unstructured, unordered, and hence it is not easy to apply convolutions directly while maintaining order invariance. A pioneering point-based method, PointNet \cite{PointNet} addresses this problem by utilizing symmetric operations to derive a feature representation for point clouds. An improved work PointNet++ \cite{PointNet++} uses a hierarchical grouping of points to capture local and global information. However, strong dependence on neighbouring points makes it suffer greatly on unseen point corruptions. A few point-based methods \cite{DGCNN} use graph-convolutions for feature extraction by utilizing local neighbourhood information. All these methods use convolutions and assume a uniform input distribution, which is not valid in real datasets containing various imperfections. Also, convolutions cannot learn long-range dependencies, limiting the quality of contextual information derived from the point cloud. Recent works \cite{Gumbel, SetT, L2G, ShapeContextNet} have shown that attention in conjunction with convolutions improve performance. Attentional ShapeContextNet \cite{ShapeContextNet} employs sequential attention blocks to derive point features, max pooled to create a global feature representation and causes loss of information. Set Transformer \cite{SetT} uses a similar idea and introduces an attention-based pooling operation but leads to over-fitting due to excess parameterization.
\begin{figure*}[t!]
    \hskip-1cm\begin{subfigure}[t]{0.4\textwidth}
        \centering
        \includegraphics[height=1.5in]{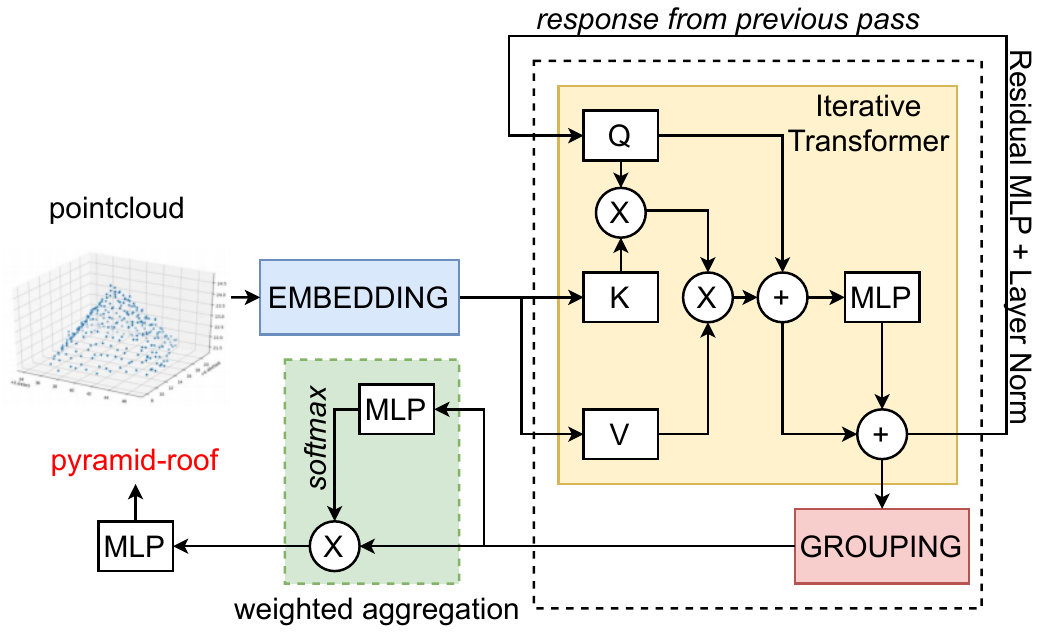}
        \caption{\scriptsize{\label{fig:TB}Point Transformer Network}}
    \end{subfigure}%
    ~ 
    \begin{subfigure}[t]{0.35\textwidth}
        \centering
        \includegraphics[height=1.2in]{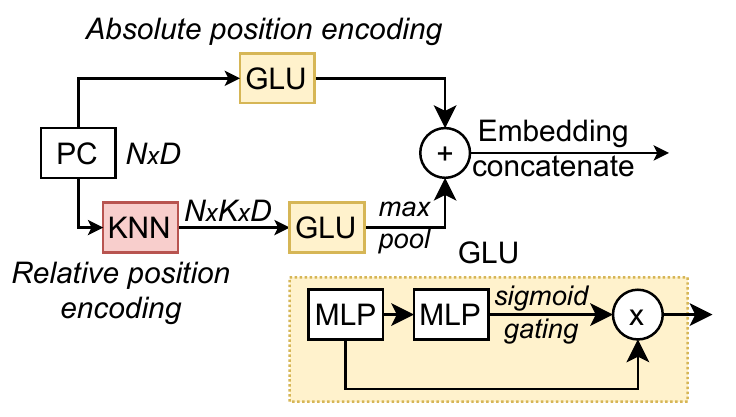}
        \caption{\scriptsize{\label{fig:EM}Embedding Module}}
    \end{subfigure}
    ~ 
     \begin{subfigure}[t]{0.3\textwidth}
       \centering
        \includegraphics[height=1.6in]{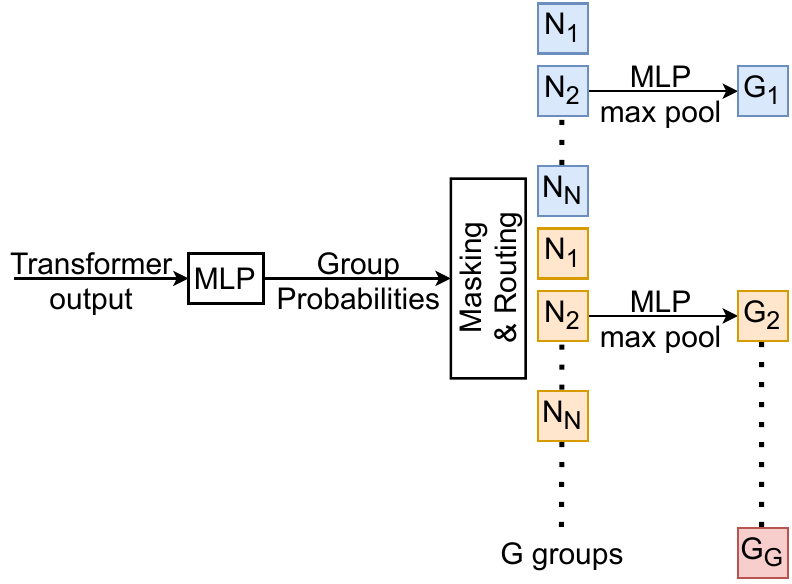}
        \caption{\scriptsize{\label{fig:GM}Grouping Module}}
    \end{subfigure}
    \caption{\small{\label{fig:PTN} Architecture of Point Transformer Network}}
\end{figure*}

This letter explores the possibility of completely replacing convolutional layers with attention blocks. Our motivation for this study is that convolutional kernels get activated at specific regions, and similarly, attention layers focus on specific regions of the point cloud. The proposed method - Point Transformer (PT) outperforms other state-of-the-art models in a real dataset - RoofN3D \cite{RoofN3D}, and to ensure a fair comparison with previous works, we evaluate the model on a standard benchmark dataset - ModelNet40. Even though most methods achieve high classification accuracy in clean, aligned datasets, they perform poorly in unseen point cloud transformations. Therefore, to determine the model's robustness to various unseen point corruptions (noise, sparsity, occlusion, etc.), it is tested on a dataset RobustPointSet \cite{RobustPointSet} and the model outperforms other methods. The key contributions of our work are as follows:
\vspace*{-0.5mm}
\begin{itemize}
    \item Novel approach to show self-attention can completely replace convolutions in point cloud processing.
    \item The proposed iterative transformer (with weight-sharing) can hierarchically learn complex information with significantly lesser parameters.
    \item A learnable grouping mechanism with a routing loss is introduced to group semantically similar points and derive region/group-wise feature vectors.
    \item In this study, we also focus on the model's robustness to unseen corruptions apart from classification performance on benchmark datasets.
\end{itemize}

\section{Related Works}
The methods for modeling building rooftops from ALS data are categorized into model-driven and data-driven techniques. Model-driven approaches fit the roof points to a given set of roof templates and select the best fit. This method is parametric, resilient to noise and missing data but requires prior information about the roof shape \cite{isprs-2018}. There are various data-driven methods like plane fitting, filtering and thresholding, segmentation-based, and learning-based, which are non-parametric and do not require prior information \cite{isprs-2018}. However, it is affected by the average point density, noise, and missing points in the point cloud \cite{isprs-2018}. Though classification and retrieval of roof point clouds are prime tasks in 3-D building modeling, there are only a few learning-based methods.  Zhang et al. \cite{Zhang} introduced a random forest classifier trained using features extracted from the point cloud and a codebook, while Castagno et al. \cite{Castagno} uses a multi-modal architecture to utilize information from satellite images and ALS data. Shajahan et al. \cite{mvcnnsa} introduced a multi-view based deep-learning approach for classifying roof point clouds. These limited number of works show that deep learning techniques for shape classification and retrieval of ALS roof point clouds are still unexplored.

We propose a novel point-based deep learning approach robust to various point corruptions for shape classification and retrieval of complete and partial roof point clouds. The remainder of this letter is structured as follows: Section \ref{sec:Methodology} describes the Methodology, followed by the Experiments and Results in section \ref{sec:Results}, and Section \ref{sec:Conclusion} concludes and suggests possible improvements.
 
\section{Methodology}
\label{sec:Methodology}
Our proposed architecture - Point Transformer (PT) as shown in Fig. \ref{fig:TB} consists of the following modules:
\begin{itemize}
    \item Embedding Module
    \item Iterative Transformer
    \item Grouping Operation and Routing Loss
    \item Weighted Group Aggregation
\end{itemize}

\subsection{Embedding Module}
The input point cloud of size N represented as XYZ coordinates contains information about the absolute position of every point. While this is quite expressive, every point can also be represented by its relative position. To avail both these information, an embedding module, as shown in Fig. \ref{fig:EM} is introduced containing two branches - one to encode the absolute position and the other to derive the relative position information. The absolute position details are encoded by feeding each point into a Gated Linear Unit (GLU) to propagate the required information selectively. In the other branch, \textit{k}-nearest neighbors are computed to derive the relative position of each point. This is then passed through another GLU layer, followed by a max-pooling operation to create a suitable vector. The point representations from each branch are then concatenated to derive the enriched point-wise embedding. To make the embedding module robust to variable point cloud size, the value of \textit{k} is reduced based on the number of points. $\textit{k} = \textit{k}_{0}*N/N_{0}$ where $\textit{k}_{0}=32$, $N_{0}=1024$ are constants. 

\subsection{Iterative Transformer}
Transformers have shown huge success in natural language processing and have recently displayed some potential in image processing. They are inherently order-invariant to a given sequence, making it ideal for point cloud processing. The transformer layer is characterized by the scalar dot product (SDP) attention operation given in Eq.\ref{eq1}. The idea behind the attention mechanism is to learn complex relationships between the elements of a sequence and focus on certain parts of it. These relationships are represented by an NxN attention matrix computed as the dot product of query (Q) and key (K) vectors. This is then multiplied with the value vector (V) to derive a rich feature representation. The Q, K, V vectors are obtained using individual Multi-Layer Perceptrons (MLP) from the input sequence. Rather than computing the attention once, a Multi-Head Attention (MHA) mechanism \cite{Attention} computes the SDP operation multiple times. This helps the transformer jointly attend to different information derived from each head. 
\begin{equation} \label{eq1}
    SDP(Q,K,V) = Softmax(QK^T/\sqrt{N})V
\end{equation}
Existing methods utilize a stack of sequential transformers to learn an element-wise representation. However, given a large sequence, multiple MHA operations become very expressive, leading to overfitting \cite{SetT}. Deviating from these methods, we introduce an Iterative Transformer (shown in Fig. \ref{fig:TB}) to recursively improve the feature representation across M passes. In the first pass, the Q, K, V vectors are derived from the point embeddings, and the output is passed through a residual MLP layer with Layer Normalization. In the subsequent passes, the Q vector is updated using the output from the previous pass, and the K, V vectors are retained as the same. It is important to note that the weight parameters are shared across the passes and help learn complex features hierarchically. The above operation ensures that at every pass, the attention layers try to selectively refine the information learnt from the previous ones while keeping the number of parameters fixed irrespective of the value of M.

\subsection{Grouping Operation}
A point cloud is characterized by multiple groups of points, together representing the complete shape. As an alternative to direct pooling of the derived point representation, we create region-wise or group-wise feature vectors using a learnable grouping operation. To split the points into groups, probability values are predicted for each point, determining the best group for it. The points are then hard-routed based on these probability values, which implies that a point can belong to only one group, and no residue of the same is passed to the others. In each group, a specific feature transformation is applied to the points using an MLP layer followed by a max-pooling operation to create a group feature vector. It is imperative to ensure that for each point, the model explores all available groups before converging to the best group for a point. Hence at times during training, the points are routed to the group corresponding to their second-highest probability. This indirectly ensures that all the points are not routed to a single group. Since the model derives different features across passes, we perform the grouping operation at every pass to derive the final shape representation.

\subsubsection{Routing Loss}
During grouping, it is also essential to ensure a sufficient number of points in every group to capture region-specific information. The routing loss (given in Eq. \ref{eq2}) is introduced to penalize the model based on the distribution of points among the groups. Here $N_g$ is the number of points routed to the $g^{th}$ group, and $N$ is the total number of points in the input point cloud. 
\begin{equation} \label{eq2}
    RoutingLoss, R_L = \sum_{g=1}^{g=G} (\frac {N_g} {N}) ^ 2
\end{equation} 
To provide an intuition, let us assume $f_1$ and $f_2$ to be the fraction of points routed to any two groups. If both these fraction of points are routed to a single group, then the value of routing loss will be higher as $(f_1+f_2) ^2  >= f_1 ^2 +f_2 ^2$ and similarly, this idea can be extended to G groups. It is important to note here that the model is still free to route any number of points to a group and that the routing loss, random routing are just measures to prevent the model from getting biased to a selective number of groups. 

\subsection{Weighted Group Aggregation}
The group aggregation operation combines the group feature vectors obtained from all passes to create a point cloud representation. To select relevant information from these groups, a weight matrix is derived using an MLP layer, normalized using a softmax operation. These weights are multiplied with the corresponding features and then added to derive the global feature vector. This operation is equivalent to $sum(softmax(ax+b)*x)$ where \textit{x} denotes the input, \textit{a} and \textit{b} represent the weights and bias terms of the MLP layer respectively.

The derived global feature vector is passed onto two MLP layers to enhance it before the final classification layer. The model is trained to minimize the smooth cross entropy loss \cite{DGCNN} along with the routing loss as mentioned above. 

\section{Experiments and Results}
\label{sec:Results}
We have conducted comprehensive experiments to validate the effectiveness of Point Transformer. The details of the experiments are given below.

\subsection{Datasets}
The performance of the model is evaluated on a large scale dataset of roof point clouds - RoofN3D  \cite{RoofN3D}. This dataset contains ALS point clouds of buildings of New York city and has 1,18,073 roof shapes split into three categories - Saddleback, Two-sided Hip, and Pyramid. The dataset is uniformly split into 95,632 train, 10,627 validation, and 11,814 test samples. Being a new approach and to ensure a fair comparison with previous works, we also evaluate the model's performance on a standard benchmark dataset - ModelNet40 \cite{ModelNet40}. It contains 3-D models from 40 categories split into 9,843 train and 2,468 test samples. We use the pre-processed dataset provided by PointNet++ \cite{PointNet++} with the standard train-test split for accurate comparison. For our robustness experiments, we test a trained model on unseen transformations present in a publicly available dataset RobustPointSet \cite{RobustPointSet}. It is based on ModelNet40 with transformations applied like Noise, Missing Part, Occlusion, Sparse, Rotation, Translation. 

\subsection{Experiment Setting}
We use the following hyper-parameters to train Point Transformer for the classification task. The model derives 128 size vectors from the embedding module, subsequently fed to the transformer block with hidden size 128 and a grouping module to create group feature vectors of size 512. For all our experiments, we choose M(number of passes) = 4 and G(number of groups) = 4. The network is trained using Adam optimizer with a batch size of 18 and an initial learning rate of 0.001, which is reduced by a factor of 0.7 every 20 epochs, and the model converges in less than 100 epochs. All experiments are run on an NVIDIA GTX 1080Ti GPU.

\subsection{Shape Classification and Retrieval}
By default, we select 1000 points uniformly sampled for the RoofN3D dataset and 1024 points using farthest point sampling for ModelNet40. While training for classification, we augment the input by randomly scaling the shape in the range of $(0.8, 1.25)$ and randomly translating in the range of $({-0.1}, 0.1)$. To evaluate the shape retrieval scores, we extract the feature vectors from the penultimate MLP layer using a trained classification model. We sort the most relevant shapes for each query by cosine distance and compute the Mean Average Precision (MAP) score.

\renewcommand{\arraystretch}{1.1}
    \begin{table}[!htb]
    \caption{Shape classification and retrieval on ModelNet40} \label{tab:1}
    \begin{subtable}{.5\linewidth}\centering 
    \caption{Shape Classification results} \label{tab:1a}
     	 \begin{tabular}{|p{1.8cm}|p{1.5cm}|} 
      		\hline
      		\textbf{\scriptsize Method} &\textbf{\scriptsize {Accuracy(\%)}}\\
      		\hline
      		\textbf{\scriptsize PointNet}  & 89.2 \scriptsize \cite{PointNet}\\
      		\hline
        	\textbf{\scriptsize ASCN} & 90.0 \scriptsize \cite{ShapeContextNet}\\ 
      		\hline
      		\textbf{\scriptsize MVCNN} & 90.1 \scriptsize \cite{mvcnn}\\
      		\hline
      		\textbf{\scriptsize Set Transformer} & 90.4 \scriptsize \cite{SetT}\\ 
        	\hline
      	   	\textbf{\scriptsize L2G} & 90.6 \scriptsize \cite{L2G}\\ 
           	\hline
      		\textbf{\scriptsize PointNet++} & 90.7 \scriptsize \cite{PointNet++}\\
      		\hline
      		\textbf{\scriptsize SO-Net[\tiny 2k]} & 90.9 \scriptsize \cite{SONet}\\
      		\hline
        	\textbf{\scriptsize PAT} & 91.7 \scriptsize \cite{Gumbel}\\ 	
        	\hline
        	\textbf{\scriptsize DGCNN} & 92.2 \scriptsize \cite{DGCNN}\\ 	
        	\hline
        	\textbf{\scriptsize PointCNN} & 92.2 \scriptsize \cite{PointCNN}\\
      		\hline
        	\textbf{\scriptsize PCNN}  & 92.3 \scriptsize \cite{DensePoint}\\
      		\hline
        	\textbf{\scriptsize PT(\tiny Ours)} & \textbf{92.5}\\ 
        	\hline
      	\end{tabular}
      \end{subtable}%
      \hfill 
       	\begin{subtable}{.5\linewidth}\centering
       	 \caption{Shape Retrieval results}\label{tab:1b}
     	 \begin{tabular}{|p{1.5cm}|p{1.2cm}|} 
      		\hline
      		\textbf{\scriptsize Method} & \textbf{\scriptsize {MAP(\%)}}\\
      		\hline
      		\textbf{\scriptsize PointNet}  & 70.5 \scriptsize \cite{DensePoint}\\
      		\hline
    		\textbf{\scriptsize ASCN} & 71.83\\
            \hline
      		\textbf{\scriptsize MVCNN}  & 79.5 \scriptsize \cite{mvcnn}\\
      		\hline
      		\textbf{\scriptsize PointCNN} & 83.8 \scriptsize \cite{DensePoint}\\
      		\hline
      		\textbf{\scriptsize DGCNN} & 85.3 \scriptsize \cite{DGCNN}\\
      		\hline
      		\textbf{\scriptsize PT(\tiny Ours)} & \textbf{88.4}\\ 
        	\hline
      	\end{tabular}
      	\end{subtable}
 \end{table}
\renewcommand{\arraystretch}{1.2}
  \begin{table}[!htb]
  \centering
  	\caption{\footnotesize{\label{tab:FullR} Classification and retrieval results on RoofN3D}}
 	 \begin{tabular}{|p{1.35cm}|p{1.2cm}|p{0.8cm}|} 
  		\hline
  		\textbf{\scriptsize Method} &\textbf{\scriptsize {Accuracy(\%)}} &\textbf{\scriptsize {MAP(\%)}}\\
  		\hline
  		\textbf{\scriptsize PointNet}  & 96.57 \scriptsize \cite{SarthakTUM} & 90.8\\
  		\hline
  		\textbf{\scriptsize PointNet++} & 97.03 & 94.3\\
  		\hline
  		\textbf{\scriptsize PointCNN} & 97.39 & 94.37\\
  		\hline
  		\textbf{\scriptsize SO-Net} & 97.78 & -\\
        \hline
  		\textbf{\scriptsize MVCNN-SA} & 97.76 \scriptsize \cite{mvcnnsa}  & 96.94\\
  		\hline
    	\textbf{\scriptsize PT (Ours)} & \textbf{97.86} & \textbf{97.34}\\ 
    	\hline
  	\end{tabular}
  \hfill 
 \end{table}

\begin{figure}[!h]
     \centering
     \includegraphics[width=0.7\linewidth, height=3.5cm]{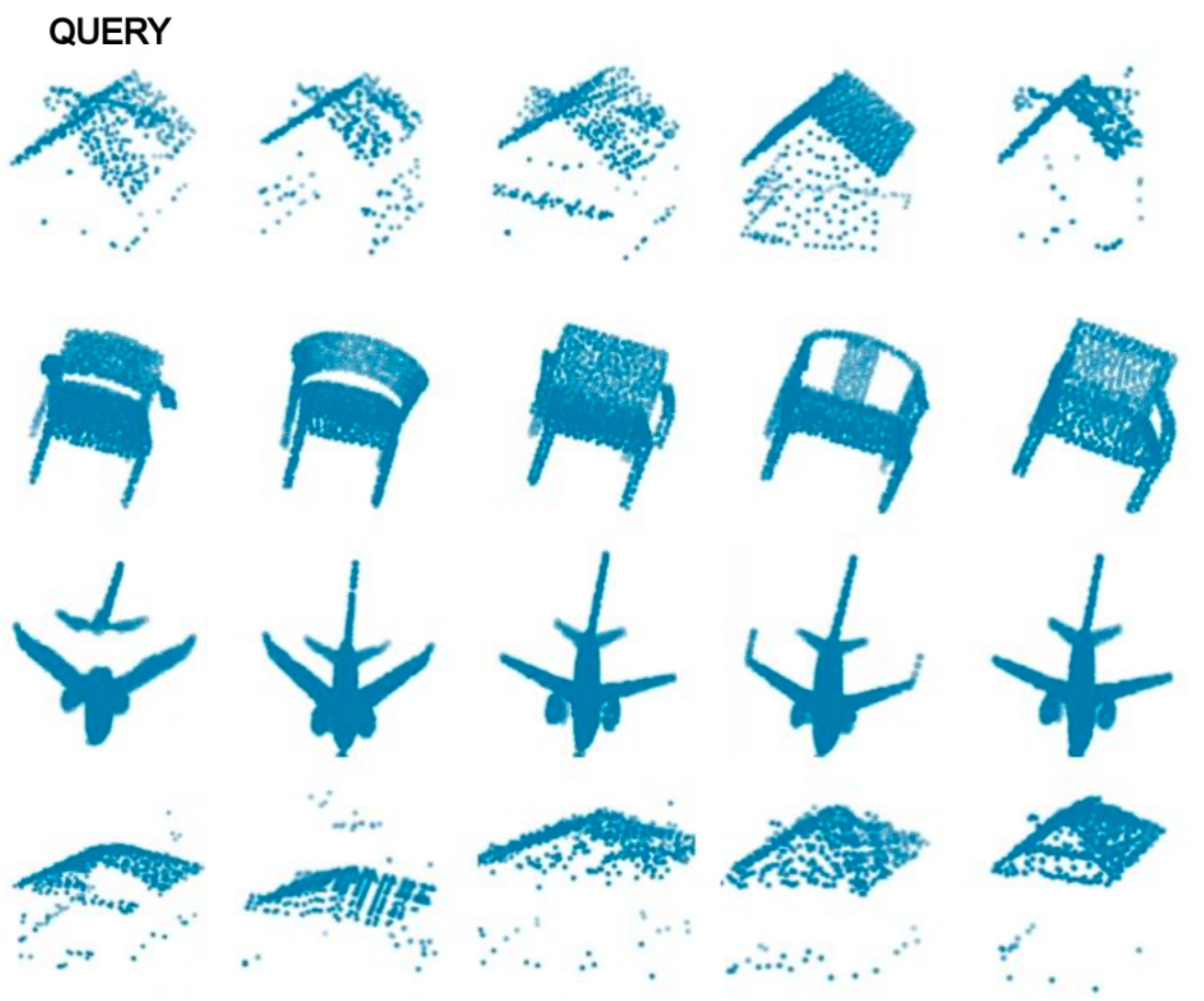}	\caption{ \scriptsize{\label{fig:retrieval}Shape retrieval using complete (top 2 rows) and partial query (bottom 2 rows)}}
\end{figure}

Tables \ref{tab:1} and \ref{tab:FullR} compare our shape classification and retrieval results with other methods on the ModelNet40 and RoofN3D datasets. Point Transformer outperforms all state-of-the-art models in the RoofN3D dataset, signifying its effectiveness in real datasets with multiple imperfections. In ModelNet40, the model performs better than \cite{PointNet++, PointCNN, SONet}, attention-based methods like \cite{L2G, ShapeContextNet}, and even graph-based methods like \cite{DGCNN}. The model also outperforms other methods in the shape retrieval task. Fig. \ref{fig:retrieval} shows samples retrieved for a given query shape. In the roof retrieval case, complex substructures present in the query shape are retained in the retrieved samples. 

\renewcommand{\arraystretch}{1.2}
  \begin{table*}[!htb]
   \caption{Results for Robustness tests in RobustPointSet (ModelNet40) and RoofN3D} \label{tab:RobTest}
  \begin{subtable}{.54\linewidth}\centering
  \caption{\small{\label{tab:4}Test for robustness in RobustPointSet}}
	 \hskip 1.0cm\begin{tabular}{ |p{1.0cm}|p{0.6cm}|p{0.6cm}|p{0.6cm}|p{0.7cm}|p{0.6cm}|p{0.6cm}|p{0.6cm}|p{0.6cm}| } 
  	    \hline
  		\textbf{\scriptsize {Method}} & \textbf{\scriptsize {Orig}} &\textbf{\scriptsize{Noise}} & \textbf{\scriptsize {Trans}} &\textbf{\scriptsize {Missing}} &\textbf{\scriptsize {Sparse}} &\textbf{\scriptsize {Rot}} &\textbf{\scriptsize {Occ}} &\textbf{\scriptsize {Avg}}\\
  		\hline
  		\textbf{\scriptsize PointNet} & 89.06 &74.72 &79.66 & 81.52 & 60.53 & 8.83 & 39.47 & 61.97\\
  		\hline
  		\textbf{\scriptsize PointNet++} & 91.47 & 14.90 & 91.07 & 50.24 & 8.85 & 12.70 & 70.23 & 48.49\\
  		\hline
  		\textbf{\scriptsize DGCNN} & 92.52 & 57.56 & 91.99 & 85.40 & 9.34 & 13.43 & 78.72 & 61.28\\
  		\hline
  		\textbf{\scriptsize PointMask} & 88.53 & 73.14 & 78.20 & 81.48 & 58.23 & 8.02 & 39.18 & 60.97\\
  		\hline
  		\textbf{\scriptsize DensePoint} & 90.96 & 53.28 & 90.72 & 84.49 & 15.52 & 12.76 & 67.67 & 59.40\\
  		\hline
  		\textbf{\scriptsize PointCNN} &  87.66 & 45.55 & 82.85 & 77.60 & 4.01 & 11.50 & 59.50 & 52.67\\
  		\hline
  		\textbf{\scriptsize PointConv} & 91.15 & 20.71 & 90.99 & 84.09 & 8.65 & 12.38 & 45.83 & 50.54\\
  		\hline
  		\textbf{\scriptsize RSCNN} & 91.77 & 48.06 & 91.29 & 85.98 & 23.18 & 11.51 & 75.61 & 61.06\\
		\hline
    	\textbf{\scriptsize \textbf{PT} (\tiny Ours)} & \textbf{92.39} & \textbf{71.52} & \textbf{92.1} & \textbf{86.62} & \textbf{15.1} &  \textbf{14.02} & \textbf{66.1} & \textbf{62.55}\\
    	\hline
  	\end{tabular}
  	\end{subtable}%
  	\hfill
  	  \begin{subtable}{.49\linewidth}
  	  \begin{center}
  	 \caption{\small{\label{tab:tab5}Test for robustness in RoofN3D}}
	 	 \begin{tabular}
	 	 	 { |p{1.0cm}|p{0.6cm}|p{0.6cm}|p{0.65cm}|p{1.0cm}|p{1.5cm}| } 
  	    \hline
  		\textbf{\scriptsize {Method}} & \textbf{\scriptsize {Orig}} &\textbf{\scriptsize{Noise (75\%)}} & \textbf{\scriptsize {Missing (50\%)}} &\textbf{\scriptsize {Uniform removal (87.5\%)}} &\textbf{\scriptsize {Missing shape retrieval (75\%) (MAP)}}\\
  		\hline
  		\textbf{\scriptsize PointNet} & 96.57 & 95.07 & 83.22 & 75.52 & 83.89\\
  		\hline
  		\textbf{\scriptsize PointNet++} & 97.03 & 75.55 & 77.73 & 70.02 & 83.45\\
  		\hline
  		\textbf{\scriptsize PointCNN} &  97.39 & 76.32 & 80.43 & 73.60 & 86.5\\
  		\hline
    	\textbf{\scriptsize \textbf{PT} (\tiny Ours)} & \textbf{97.86} & \textbf{97.6} & \textbf{92.3} & \textbf{76.92} & \textbf{90.77}\\
    	\hline
    \end{tabular}
    \vspace{-2mm}
    \end{center}
    \begin{tablenotes}
    \item \textit{\scriptsize{Table \ref{tab:4}: Orig-Original, Trans-Translation, Rot-Rotation, Occ-Occlusion, Avg-Average Accuracy. Table \ref{tab:tab5}: Orig-Original, Noise(75\%)-gaussian noise added to 75\% points, Missing(50\%)-50\% points are removed from a single region, Uniform removal(87.5\%)-87.5\% points are uniformly removed}}
    \end{tablenotes}
  	\end{subtable}
 \end{table*}  

\subsection{Test for Robustness}
\label{subsec:Robust}
Most existing methods primarily focus on improving classification accuracy on clean, aligned datasets. It is equally important to test the model's robustness on unseen transformations to prove its applicability for real datasets. For this experiment, we follow the procedure given in RobustPointSet by training the model on 2048 points without any augmentation and test it for various unseen corruptions. Table \ref{tab:RobTest} shows results for various models on these unseen test sets, and it is quite evident that our proposed method outperforms other techniques. Among the other methods, PointNet seems to be a robust model even though it doesn't achieve state-of-the-art performance in the classification task. As seen from the Table \ref{tab:4}, most methods fail significantly in cases like Sparse, Noise, Rotation. Methods that rely on local information suffer greatly since the notion of neighbourhood is changed drastically on corruption. During rotation, 3D coordinate positions change and hence affects the performance of all models. We extend these experiments to the RoofN3D dataset, and the results are shown in Table \ref{tab:tab5}. To test for robustness in the shape retrieval task, partial point clouds are used as the query shape, and corresponding closest matching full shapes are retrieved. This further validates the expressivity of our method in spite of being given incomplete input shapes.

\begin{figure}[t!]
    \begin{subfigure}[t]{0.23\textwidth}
        \includegraphics[width=\textwidth, height =3.2cm]{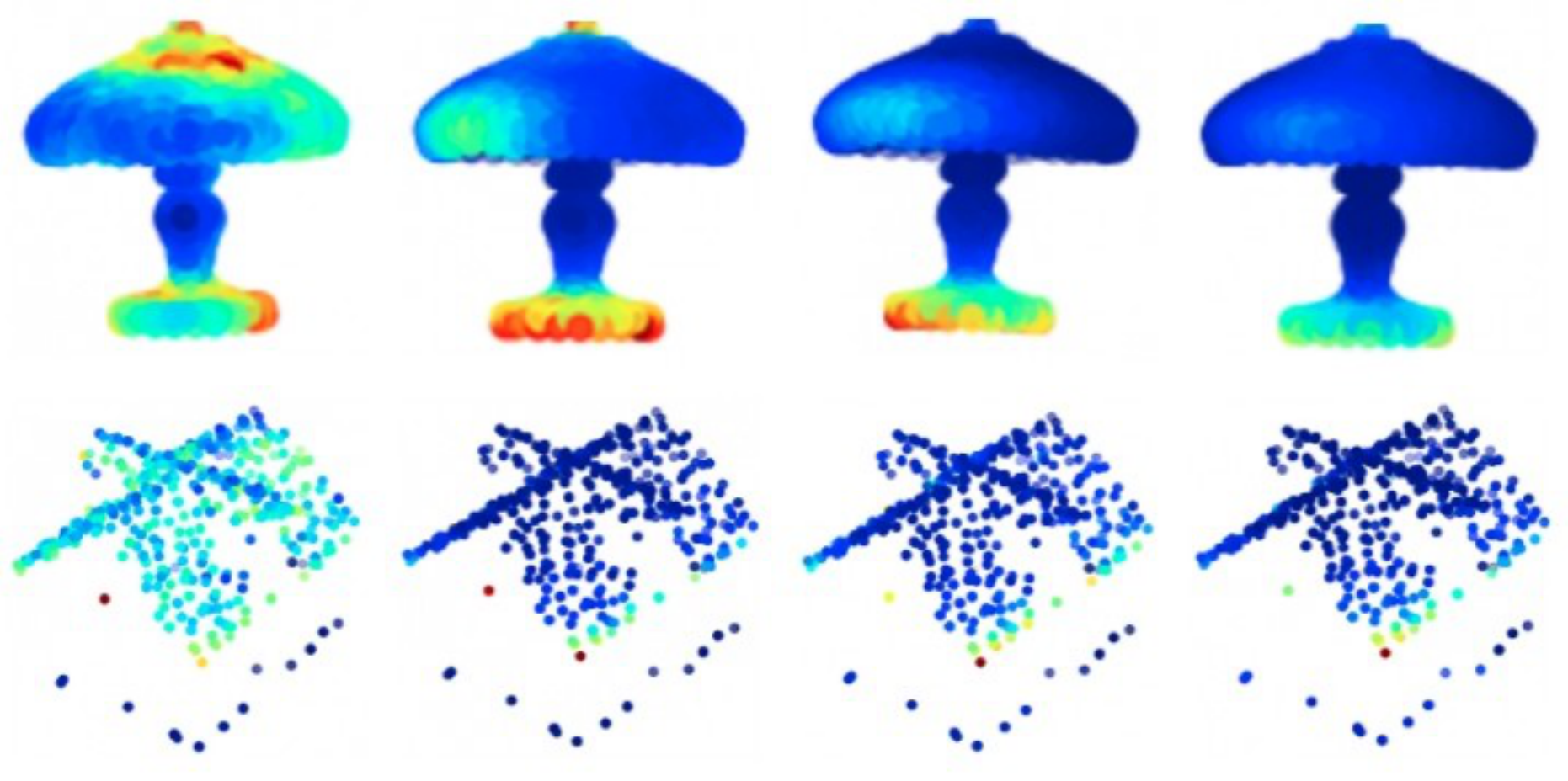}
        \caption{\scriptsize{\label{fig:avg_atn} Average attention across passes (Red-High, Blue-Low)}}
    \end{subfigure}%
    ~ ~ ~ ~ ~
    \begin{subfigure}[t]{0.23\textwidth}
        \includegraphics[width=\textwidth,height =3.2cm]{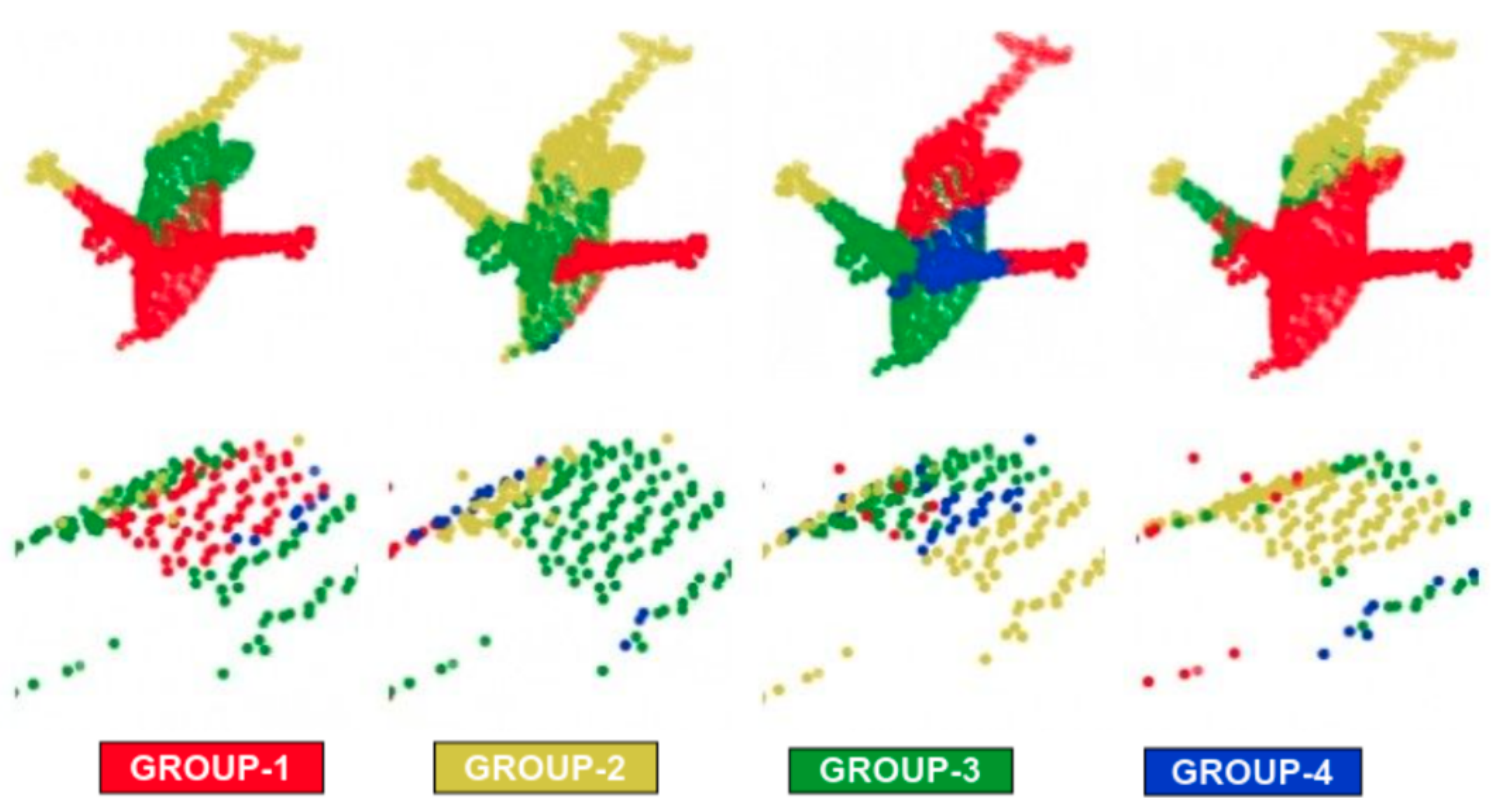}
        \caption{\scriptsize{\label{fig:gp_atn}Group formation across passes (each color indicates a group)}}
    \end{subfigure}
     \caption{\scriptsize{\label{fig:vis}Visualising the representations learnt by the Model}}
\end{figure}

\subsection{Visualising the Model}
\begin{figure}[t!]
    \hskip -0.5cm\begin{subfigure}[t]{0.23\textwidth}
        \includegraphics[width=\textwidth, height=4cm]{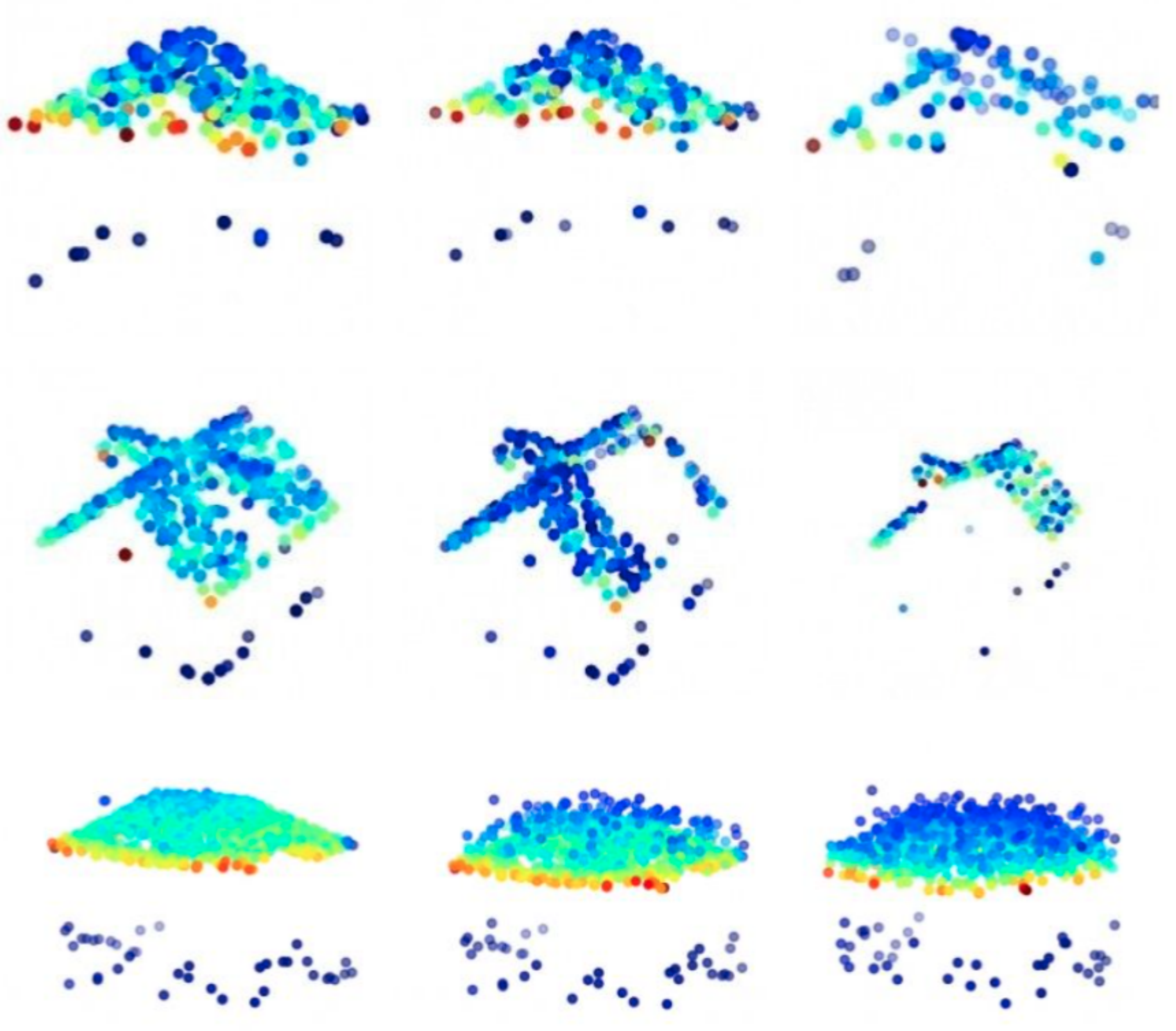}
        \caption{\scriptsize{RoofN3D}}
    \end{subfigure}%
    ~ ~ ~ ~ ~
    \begin{subfigure}[t]{0.23\textwidth}
        \includegraphics[width=\textwidth, height=4cm]{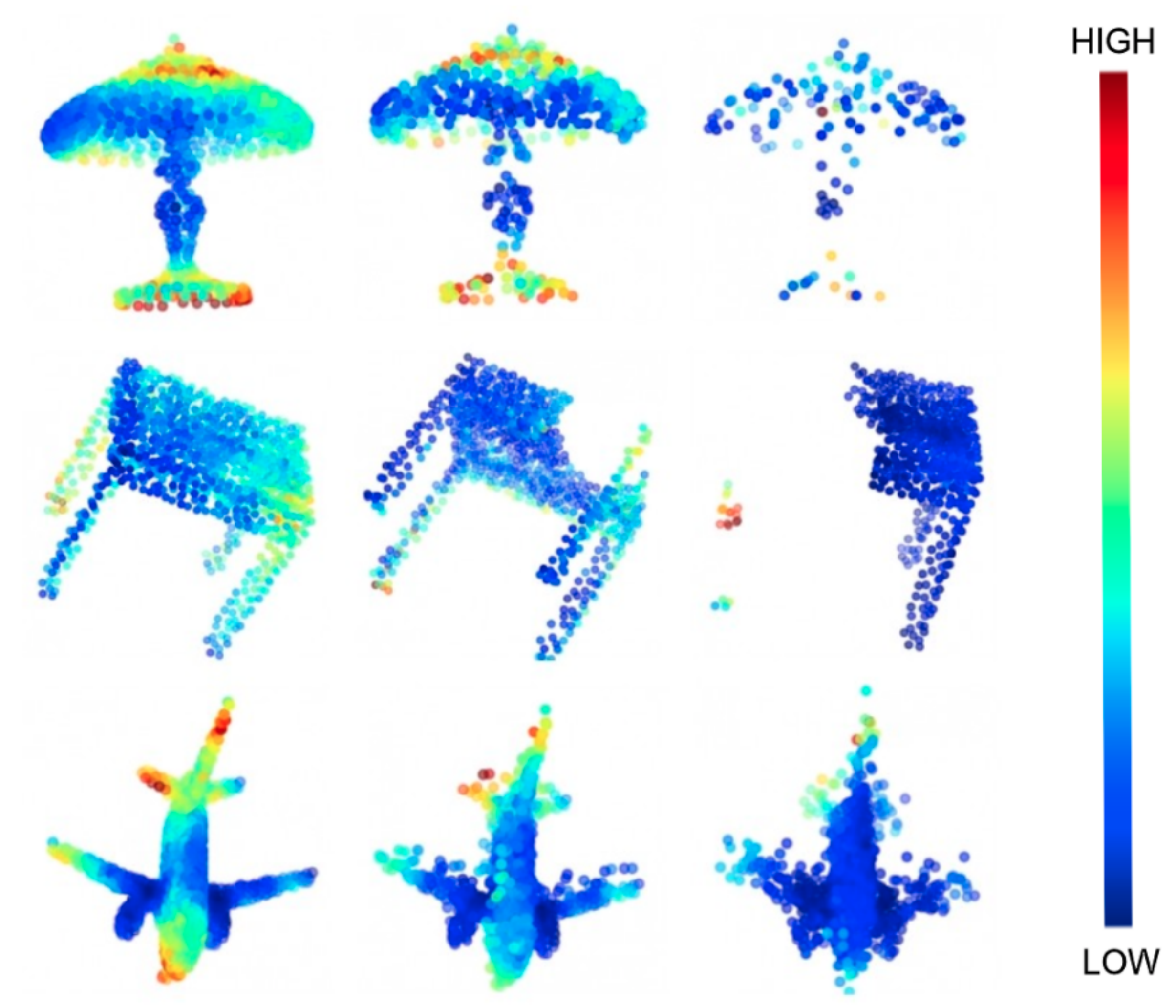}
        \caption{\scriptsize{ModelNet40}}
    \end{subfigure}
    \caption{\scriptsize{\label{fig:rob}Attention Maps (Red-High, Blue-Low) on various levels of Point Corruptions: reduced point density, partial shape removal, noise addition. (from top to bottom)}}
\end{figure}

To better understand and validate the operations in our method, we visualize the average attention maps and group formation. We compute the average attention by taking the mean across heads and for each point. As seen from Fig. \ref{fig:avg_atn}, the model distributes its attention across all points in the initial pass but later attends to specific regions, as it tries to iteratively improve the feature representation. Semantically similar points are grouped together describing a particular shape in the point cloud, as shown in Fig. \ref{fig:gp_atn} where each group is given a different color. We also visualize the variation in average attention on increasing point corruptions, as shown in Fig. \ref{fig:rob} and the model has focused on similar regions in all cases. An interesting observation is that the model can identify noisy points and not pay attention to them.

 \renewcommand{\arraystretch}{1.2}
  \begin{table}[!h]
  	\centering
 	 \caption{\small{\label{tab:TC} Model Parameters, Size and Forward Pass time}}
	 \begin{tabular}{|p{2cm}|p{1.25cm}|p{1.25cm}|p{1.2cm}|}
  		\hline
  		\textbf{\scriptsize Method} & \textbf{\scriptsize Params(M)} & \textbf{\scriptsize {Size (MB)}} &\textbf{\scriptsize {Time (ms)}}\\
  		\hline
  		\textbf{\scriptsize MVCNN(12 views)} & 128.93 & 515.7 & 55.6\\
        \hline
		\textbf{\scriptsize MVCNN-SA} & 15.34 & 61.5 & 27.2\\
        \hline
  		\textbf{\scriptsize PointNet} & 3.5 & 40 & 25.3\\
  		\hline
  		\textbf{\scriptsize PointNet++ (MSG)} & 1.48 & 12 & 163.2\\
  		\hline
  		\textbf{\scriptsize PointCNN} & 0.6 \cite{PointCNN} & 6.9 \cite{PointCNN}& -\\
  		\hline
    	\textbf{\scriptsize PT} & \textbf{1.03} & \textbf{4.1} & \textbf{10.9}\\ 
    	\hline
  	\end{tabular}
 \end{table} 
 
\subsection{Time and Space Complexity}
\label{sec:Time}
While view-based methods achieve high performance, they are very computationally expensive and less space-efficient than point-based methods, as shown in Table. \ref{tab:TC}. Among the point-based methods, Point Transformer is a much smaller model with only 1.03 M parameters compared to other methods like PointNet, PointNet++ having 3.5M and 1.48M parameters, respectively. Apart from the number of parameters, our method has significantly lower computational time as it replaces all convolutions with fully connected layers. This proves the efficiency of our model without compromising on performance. 
\vspace{-2ex}

\section{Conclusion}
\label{sec:Conclusion}
Inspired by the success of Transformers in Natural Language Processing and recently in image processing, we introduce a stand-alone self attention model, Point Transformer for efficient point cloud representation. We propose a novel iterative transformer, learnable grouping operation which makes our method more effective. Detailed experiments validate the effectiveness of our method in shape classification and retrieval tasks. Further, we also focus on the robustness of our model on unseen input transformations and it outperforms other state-of-the-art techniques. The model is also highly time and memory efficient when compared to other methods. We hope to inspire future work to investigate the properties of point transformer networks and extend it for various 3D pointcloud processing tasks. 



\begin{thebibliography}{10}
\providecommand{\url}[1]{#1}
\csname url@samestyle\endcsname
\providecommand{\newblock}{\relax}
\providecommand{\bibinfo}[2]{#2}
\providecommand{\BIBentrySTDinterwordspacing}{\spaceskip=0pt\relax}
\providecommand{\BIBentryALTinterwordstretchfactor}{4}
\providecommand{\BIBentryALTinterwordspacing}{\spaceskip=\fontdimen2\font plus
\BIBentryALTinterwordstretchfactor\fontdimen3\font minus
  \fontdimen4\font\relax}
\providecommand{\BIBforeignlanguage}[2]{{%
\expandafter\ifx\csname l@#1\endcsname\relax
\typeout{** WARNING: IEEEtran.bst: No hyphenation pattern has been}%
\typeout{** loaded for the language `#1'. Using the pattern for}%
\typeout{** the default language instead.}%
\else
\language=\csname l@#1\endcsname
\fi
#2}}
\providecommand{\BIBdecl}{\relax}
\BIBdecl

\bibitem{ModelNet40}
{Zhirong Wu}, S.~{Song}, A.~{Khosla}, {Fisher Yu}, {Linguang Zhang}, {Xiaoou
  Tang}, and J.~{Xiao}, ``3d shapenets: A deep representation for volumetric
  shapes,'' in \emph{CVPR}, 2015, pp. 1912--1920.

\bibitem{ShapeNet}
A.~X. Chang, T.~Funkhouser, L.~Guibas, P.~Hanrahan, Q.~Huang, Z.~Li,
  S.~Savarese, M.~Savva, S.~Song, H.~Su, J.~Xiao, L.~Yi, and F.~Yu,
  ``{ShapeNet: An Information-Rich 3D Model Repository},'' Stanford
  University-Princeton University-Toyota Technological Institute at Chicago,
  Tech. Rep. arXiv:1512.03012 [cs.GR], 2015.

\bibitem{mvcnn}
H.~Su, S.~Maji, E.~Kalogerakis, and E.~G. Learned{-}Miller, ``Multi-view
  convolutional neural networks for 3d shape recognition,'' in \emph{ICCV},
  2015.

\bibitem{Voxel}
C.~R. {Qi}, H.~{Su}, M.~{Nießner}, A.~{Dai}, M.~{Yan}, and L.~J. {Guibas},
  ``Volumetric and multi-view cnns for object classification on 3d data,'' in
  \emph{2016 IEEE CVPR}, 2016, pp. 5648--5656.

\bibitem{PointNet}
\BIBentryALTinterwordspacing
C.~R. Qi, H.~Su, K.~Mo, and L.~J. Guibas, ``Pointnet: Deep learning on point
  sets for 3d classification and segmentation,'' \emph{CoRR}, vol.
  abs/1612.00593, 2016. [Online]. Available:
  \url{http://arxiv.org/abs/1612.00593}
\BIBentrySTDinterwordspacing

\bibitem{PointNet++}
C.~R. Qi, L.~Yi, H.~Su, and L.~J. Guibas, ``Pointnet++: Deep hierarchical
  feature learning on point sets in a metric space,'' in \emph{Advances in
  Neural Information Processing Systems 30}.\hskip 1em plus 0.5em minus
  0.4em\relax Curran Associates, Inc., 2017, pp. 5099--5108.

\bibitem{PointCNN}
\BIBentryALTinterwordspacing
Y.~Li, R.~Bu, M.~Sun, and B.~Chen, ``Pointcnn,'' \emph{CoRR}, vol.
  abs/1801.07791, 2018. [Online]. Available:
  \url{http://arxiv.org/abs/1801.07791}
\BIBentrySTDinterwordspacing

\bibitem{SONet}
\BIBentryALTinterwordspacing
J.~Li, B.~M. Chen, and G.~H. Lee, ``So-net: Self-organizing network for point
  cloud analysis,'' \emph{CoRR}, vol. abs/1803.04249, 2018. [Online].
  Available: \url{http://arxiv.org/abs/1803.04249}
\BIBentrySTDinterwordspacing

\bibitem{DensePoint}
Y.~Liu, B.~Fan, G.~Meng, J.~Lu, S.~Xiang, and C.~Pan, ``Densepoint: Learning
  densely contextual representation for efficient point cloud processing,'' in
  \emph{ICCV}, October 2019.

\bibitem{DGCNN}
Y.~Wang, Y.~Sun, Z.~Liu, S.~E. Sarma, M.~M. Bronstein, and J.~M. Solomon,
  ``Dynamic graph cnn for learning on point clouds,'' \emph{ACM Transactions on
  Graphics (TOG)}, 2019.

\bibitem{Gumbel}
\BIBentryALTinterwordspacing
J.~Yang, Q.~Zhang, B.~Ni, L.~Li, J.~Liu, M.~Zhou, and Q.~Tian, ``Modeling point
  clouds with self-attention and gumbel subset sampling,'' in
  \emph{CVPR}.\hskip 1em plus 0.5em minus 0.4em\relax Computer Vision
  Foundation / {IEEE}, 2019. [Online]. Available:
  \url{http://openaccess.thecvf.com/content\_CVPR\_2019/html/Yang\_Modeling\_Point\_Clouds\_With\_Self-Attention\_and\_Gumbel\_Subset\_Sampling\_CVPR\_2019\_paper.html}
\BIBentrySTDinterwordspacing

\bibitem{SetT}
J.~Lee, Y.~Lee, J.~Kim, A.~R. Kosiorek, S.~Choi, and Y.~W. Teh, ``Pytorch
  implementation of set transformers,''
  \url{https://github.com/juho-lee/set_transformer/issues/3}, 2019.

\bibitem{L2G}
\BIBentryALTinterwordspacing
X.~Liu, Z.~Han, X.~Wen, Y.~Liu, and M.~Zwicker, ``{L2G} auto-encoder:
  Understanding point clouds by local-to-global reconstruction with
  hierarchical self-attention,'' in \emph{ACM MM 2019}.\hskip 1em plus 0.5em
  minus 0.4em\relax {ACM}, 2019, pp. 989--997. [Online]. Available:
  \url{https://doi.org/10.1145/3343031.3350960}
\BIBentrySTDinterwordspacing

\bibitem{ShapeContextNet}
S.~{Xie}, S.~{Liu}, Z.~{Chen}, and Z.~{Tu}, ``Attentional shapecontextnet for
  point cloud recognition,'' in \emph{2018 IEEE/CVF Conference on Computer
  Vision and Pattern Recognition}, 2018, pp. 4606--4615.

\bibitem{RoofN3D}
\BIBentryALTinterwordspacing
A.~Wichmann, A.~Agoub, and M.~Kada, ``Roofn3d: Deep learning training data for
  3d building reconstruction,'' \emph{ISPRS Archives}, vol. XLII-2, pp.
  1191--1198, 2018. [Online]. Available:
  \url{https://www.int-arch-photogramm-remote-sens-spatial-inf-sci.net/XLII-2/1191/2018/}
\BIBentrySTDinterwordspacing

\bibitem{RobustPointSet}
S.~A. Taghanaki, J.~Luo, R.~Zhang, Y.~Wang, P.~K. Jayaraman, and K.~M.
  Jatavallabhula, ``Robustpointset: A dataset for benchmarking robustness of
  point cloud classifiers,'' 2020.

\bibitem{isprs-2018}
\BIBentryALTinterwordspacing
M.~Gkeli and C.~Ioannidis, ``Automatic 3d reconstruction of buildings roof tops
  in densely urbanized areas,'' \emph{ISPRS}, vol. XLII-4/W10, pp. 47--54,
  2018. [Online]. Available:
  \url{https://www.int-arch-photogramm-remote-sens-spatial-inf-sci.net/XLII-4-W10/47/2018/}
\BIBentrySTDinterwordspacing

\bibitem{Zhang}
\BIBentryALTinterwordspacing
X.~Zhang, A.~Zang, G.~Agam, and X.~Chen, ``Learning from synthetic models for
  roof style classification in point clouds,'' in \emph{Proceedings of the 22Nd
  ACM SIGSPATIAL International Conference on Advances in Geographic Information
  Systems}, ser. SIGSPATIAL '14.\hskip 1em plus 0.5em minus 0.4em\relax New
  York, NY, USA: ACM, 2014, pp. 263--270. [Online]. Available:
  \url{http://doi.acm.org/10.1145/2666310.2666407}
\BIBentrySTDinterwordspacing

\bibitem{Castagno}
J.~Castagno and E.~Atkins, ``Roof shape classification from lidar and satellite
  image data fusion using supervised learning,'' \emph{Sensors}, vol.~18, p.
  3960, 11 2018.

\bibitem{mvcnnsa}
D.~A. {Shajahan}, V.~{Nayel}, and R.~{Muthuganapathy}, ``Roof classification
  from 3-d lidar point clouds using multiview cnn with self-attention,''
  \emph{IEEE GRSL}, vol.~17, no.~8, pp. 1465--1469, 2020.

\bibitem{Attention}
\BIBentryALTinterwordspacing
A.~Vaswani, N.~Shazeer, N.~Parmar, J.~Uszkoreit, L.~Jones, A.~N. Gomez, L.~u.
  Kaiser, and I.~Polosukhin, ``Attention is all you need,'' in \emph{Advances
  in Neural Information Processing Systems 30}.\hskip 1em plus 0.5em minus
  0.4em\relax Curran Associates, Inc., 2017, pp. 5998--6008. [Online].
  Available:
  \url{http://papers.nips.cc/paper/7181-attention-is-all-you-need.pdf}
\BIBentrySTDinterwordspacing

\bibitem{SarthakTUM}
S.~Guptha and R.~Bohare, ``Project title,''
  \url{https://github.com/sarthakTUM/roofn3d}, 2019.

\end{thebibliography}
\end{document}